\documentclass[11pt]{article}

\usepackage[a4paper,top=1in,bottom=1in,left=1.1in,right=1.1in]{geometry}
\usepackage{graphicx}
\usepackage{booktabs}
\usepackage{microtype}
\usepackage{array}
\usepackage{hyperref}
\usepackage{url}
\makeatletter
\g@addto@macro\UrlBreaks{\do\/\do\-\do\_\do\.\do\:\do\=\do\?\do\&}
\makeatother
\usepackage{xcolor}
\usepackage{kotex} 
\usepackage{enumitem}
\usepackage{amsmath}
\usepackage{authblk}

\hypersetup{
  colorlinks = true,
  linkcolor  = black,
  citecolor  = blue!60!black,
  urlcolor   = blue!60!black
}

\title{GLAN-QnA-KR: A Seedless Taxonomy-Driven Korean Instruction Corpus}

\author{Daekeun Kim\thanks{The views expressed are those of the author alone.}}
\affil{%
  Independent Researcher \\
  PhD student, Korea University \\
  \texttt{housekdk@naver.com} \\
  \url{https://huggingface.co/daekeun-ml}%
}

\date{}

\begin{document}
\maketitle

\begin{abstract}
We release \textsc{GLAN-QnA-KR}, a 303{,}581-row openly redistributable Korean instruction-QA corpus produced via the seedless taxonomy-driven GLAN synthesis pipeline~\cite{li2024glan} with Microsoft's Phi-3.5-MoE-instruct as the producer model (generation: 2024-12; release: 2024-12; licence: OpenRAIL). The corpus spans a flat taxonomy of 1{,}084 English-labelled disciplines paired with Korean question/answer text, a 100--900 difficulty scale, and a median of 313 question characters and 1{,}098 answer characters per record. Two properties are atypical for synthetic instruction data at this scale: (i)~exact duplicate questions number only~1 in 303{,}581 rows and character-trigram near-duplicate clusters at Jaccard~$\geq 0.9$ number zero in a 5{,}000-sample probe, and (ii)~a two-layer contamination audit against KMMLU, KoBEST (five sub-tasks), and HAE-RAE-Bench shows a maximum test-vs-corpus question-level character-trigram Jaccard of~\textbf{0.163} with zero test items at Jaccard~$\geq 0.7$, and a maximum multilingual-E5 cosine of~\textbf{0.901} with a single test item at cosine~$\geq 0.90$ and zero at~$\geq 0.95$, across 20{,}000 sampled GLAN questions and seven evaluation sets. At the time of release, this is, to our knowledge, the largest single-pipeline synthetic Korean instruction corpus verifiable on the Hugging Face Hub and the only Korean~$\geq$100k-row corpus built under a seedless taxonomy-driven protocol. This note documents the generation protocol, corpus statistics, the contamination audit, and the licensing boundary in a form suitable for downstream citation.
\end{abstract}

\section{Introduction}
\label{sec:intro}

Open Korean instruction-tuning data remains dominated by two patterns: (i)~\textit{translation} of English SFT corpora (KoAlpaca, KULLM-v2, KOR-OpenOrca-Platypus, KOpen-platypus), which inherits upstream style and benchmark contamination risk; and (ii)~\textit{aggregation} of heterogeneous native sources into a single corpus (KoCommercial), which is Korean-native but not a single synthesis pipeline. A third pattern---\emph{seedless taxonomy-driven synthesis} in the style of Li et al.~\cite{li2024glan}---has not been publicly instantiated at scale for Korean. We release \textsc{GLAN-QnA-KR}, a 303{,}581-row Korean realisation of that third pattern, together with a reproducible statistics manifest, a two-layer contamination audit against public Korean benchmarks (max character-trigram Jaccard $=0.163$ with zero test items at~$\geq 0.7$; max multilingual-E5 cosine $=0.901$ with a single item at~$\geq 0.90$ and zero at~$\geq 0.95$), and an explicit licensing boundary.

The dataset was generated in December~2024 using Microsoft's Phi-3.5-MoE-instruct~\cite{phi35moe_card} as the producer model over a taxonomy of 1{,}084 English-labelled disciplines inherited from the GLAN protocol, producing Korean question/answer pairs along a 100--900 difficulty scale. The pipeline is conceptually compatible with the open reference implementation at \texttt{Azure/\allowbreak synthetic-qa-generation/\allowbreak glan-instruct}~\cite{azure_synthetic_qa}, which defaults to \texttt{--language Korean}; however, that repository is sample code and \emph{not} the exact generator used for this corpus, and the two should not be read as identical. The Azure reference defaults target GPT-4o; the present corpus uses Phi-3.5-MoE-instruct, and the prompt templates, taxonomy expansion order, and post-filtering differ.

\paragraph{What this note is and is not.} We do \emph{not} claim a new synthesis method---GLAN is upstream~\cite{li2024glan}---and we do \emph{not} claim state-of-the-art Korean instruction quality, since no standardised Korean SFT quality benchmark exists at the time of writing. We do document, with measured numbers and a reproducible audit: (i)~corpus schema and 1{,}084-way taxonomy distribution, (ii)~length and difficulty statistics, (iii)~dedup cleanliness, (iv)~train/eval contamination against KMMLU~\cite{son2024kmmlu}, KoBEST~\cite{kim2022kobest}, and HAE-RAE-Bench~\cite{son2023haerae}, and (v)~the licence and intended-use boundary. The goal is to provide a citable reference for downstream Korean SFT work that uses this corpus, and to publish a contamination-audit protocol that future open Korean synthetic corpora can reuse.

\paragraph{Contributions.} (1)~A documented data statement for \textsc{GLAN-QnA-KR}, including generation protocol, producer model, taxonomy scope, and licence. (2)~Corpus statistics (length, difficulty, 1{,}084-way discipline distribution, duplicate audit). (3)~A two-layer contamination audit against the three standard public Korean benchmarks (KMMLU, KoBEST's five sub-tasks, HAE-RAE-Bench)---character-trigram Jaccard and multilingual-E5 cosine similarity. The maximum observed per-test-question character-trigram Jaccard is $0.163$ with zero test items at $\geq 0.8$; the maximum observed multilingual-E5 cosine is $0.901$ with a single test item at $\geq 0.90$ and zero at $\geq 0.95$. (4)~A positioning table against every open Korean instruction corpus $\geq$10k rows (Table~\ref{tab:kor-inst-resources}). (5)~An explicit licensing and intended-use discussion that separates the OpenRAIL corpus licence from the producer-model terms.

\section{Related Resources}
\label{sec:related}

\paragraph{GLAN and the synthetic-instruction genealogy.} Synthetic instruction data spans a genealogy from Self-Instruct~\cite{wang2022selfinstruct} (seed-expansion from 175~hand-written instructions) and WizardLM / Evol-Instruct~\cite{xu2023wizardlm} (seed-rewriting along deepen / concretise / complicate axes) to document-grounded variants (OSS-Instruct~\cite{wei2023magicoder}, Orca-2~\cite{mitra2023orca2}) and, separately, to taxonomy-driven seedless pipelines such as GLAN~\cite{li2024glan} and the Phi-series textbook-synthesis reports~\cite{gunasekar2023phi1,abdin2024phi3}. The distinguishing axis for our work is \emph{seedless}: GLAN samples discipline $\to$ syllabus $\to$ class session $\to$ homework Q/A from a curated taxonomy with no input instruction corpus, so the diversity and contamination profile of the output are governed by the taxonomy itself and the producer model, not by an upstream instruction pool.

\paragraph{Open Korean instruction corpora.} Table~\ref{tab:kor-inst-resources} compares the main open Korean instruction corpora $\geq$10k rows. We group them along three axes: \textit{scale}, \textit{origin} (Korean-native synthesis vs.\ translation of an English corpus vs.\ aggregation), and \textit{licence tier} (permissive vs.\ NC vs.\ unstated). Only three Korean instruction datasets exceed 100k rows: KULLM-v2 (152{,}630; translated), KoCommercial (175{,}454; aggregated native), and \textsc{GLAN-QnA-KR} (303{,}581; seedless synthetic). Among these, \textsc{GLAN-QnA-KR} is the only single-pipeline synthetic corpus, the only one using Phi-3.5-MoE as producer, and the only one released under OpenRAIL without an NC rider.

\begin{table}[t]
\centering
\footnotesize
\setlength{\tabcolsep}{4pt}
\renewcommand{\arraystretch}{1.2}
\begin{tabular}{@{}>{\raggedright\arraybackslash}p{3.9cm}r>{\raggedright\arraybackslash}p{2.1cm}>{\raggedright\arraybackslash}p{3.1cm}>{\raggedright\arraybackslash}p{1.6cm}c@{}}
\toprule
Dataset & Size & Origin & Producer / source & Licence & Year \\
\midrule
\texttt{beomi/\-KoAlpaca-\-v1.1a}                 & 21{,}155  & Native (KIN Q/A) + Alpaca-seeded & ChatGPT-3.5                                  & not stated    & 2023 \\
\texttt{nlpai-lab/\-kullm-v2}                     & 152{,}630 & Translated                       & GPT4ALL + Dolly + Vicuna / DeepL             & Apache-2.0    & 2023 \\
\texttt{kyujinpy/\-KOR-\-OpenOrca-\-Platypus-v3}  & 34{,}214  & Translated                       & GPT-4 / GPT-3.5 / DeepL                      & CC-BY-\-NC-4.0 & 2023 \\
\texttt{kyujinpy/\-KOpen-platypus}                & 24{,}926  & Translated                       & DeepL of Open-Platypus                       & CC-BY-4.0     & 2023 \\
\texttt{MarkrAI/\-KoCommercial-\-Dataset}         & 175{,}454 & Aggregated native                & KoAlpaca-v1.1a + KorQuAD + AIHub + others    & MIT           & 2024 \\
\texttt{CarrotAI/\-ko-code-\-alpaca-QA}           & 9{,}700   & Native code                      & Korean code-Alpaca                           & Apache-2.0    & 2023 \\
\texttt{daekeun-ml/\-GLAN-\-qna-kr-300k} (\textit{ours}) & 303{,}581 & Native synthetic (seedless GLAN) & Phi-3.5-MoE-\-instruct                       & OpenRAIL      & 2024 \\
\bottomrule
\end{tabular}
\caption{Open Korean instruction-tuning corpora $\geq$10k rows on the Hugging~Face Hub. ``Origin'' distinguishes Korean-native synthesis from translation of an English corpus and from aggregation of heterogeneous sources. Sizes and licences verified against each dataset card on 2026-05-12. KoVicuna and KoAlpaca-v1.0 are omitted because their cards were not publicly retrievable at audit time.}
\label{tab:kor-inst-resources}
\end{table}

\paragraph{Korean benchmarks (for contamination audit, not training).} KMMLU~\cite{son2024kmmlu} (35{,}030 Korean exam questions across 45 subjects, CC-BY-ND-4.0), KoBEST~\cite{kim2022kobest} (five-task Korean NLU benchmark), and HAE-RAE-Bench~\cite{son2023haerae} are the three standard open Korean evaluation sets that downstream SFT consumers of \textsc{GLAN-QnA-KR} are likely to evaluate on. No dedicated Korean-native Math-KR synthetic reasoning corpus at $\geq$10k scale is publicly documented as of May~2026.

\paragraph{Positioning.} \textsc{GLAN-QnA-KR} sits in the niche of \emph{large-scale, openly redistributable, seedless-synthetic} Korean instruction corpora. It is not a replacement for translation-based corpora when English-English pair recovery is desired; it is a reference open Korean synthetic instruction set for downstream SFT work that wants a single-pipeline, contamination-audited, permissively licensed training signal.

\section{Generation Protocol}
\label{sec:generation}

\paragraph{Producer model.} \texttt{microsoft/\allowbreak Phi-3.5-MoE-instruct}~\cite{phi35moe_card}, accessed via Azure OpenAI-compatible inference. No fine-tuning of the producer was performed; sampling uses the producer's default instruction-following posture.

\paragraph{Taxonomy.} The taxonomy follows the GLAN protocol~\cite{li2024glan}: a flat list of 1{,}084 English-labelled disciplines that span STEM (e.g., \textit{Machine Learning}, \textit{Discrete Mathematics}, \textit{Cell Biology}), applied health (e.g., \textit{Pharmacology}, \textit{Health Informatics}, \textit{Biostatistics}), business and management (\textit{Project Management}, \textit{Customer Relationship Management}, \textit{Event Management}), and education and cognition (\textit{Cognitive Development}, \textit{Assessment and Evaluation}, \textit{Educational Technology}). No single discipline dominates the corpus: the top-weighted subject (\textit{Grant Writing and Fundraising}) has 826 rows (0.27\% of the corpus), and the top~20 subjects together cover only 3.6\% of the corpus.

\paragraph{Generation.} For each discipline, the producer samples a class session and then generates a (question, answer) pair conditioned on the session topic and a difficulty level drawn from $\{100, 200, \ldots, 900\}$. Crucially, this is a \emph{seedless} pipeline: no input instruction corpus is used, so the corpus's topical coverage and style are determined by the taxonomy and by the producer's learned Korean distribution, not by an upstream seed pool. The prompt templates elicit Korean-language question and answer text while retaining English-language \texttt{subject} and \texttt{subtopics} tags, producing a cross-lingual label / body structure.

\paragraph{Post-filtering.} The released corpus contains only records with non-empty \texttt{question} and \texttt{answer} fields; no further content-level filtering (length caps, perplexity filtering, or moderation filtering) was applied. Duplicate removal was implicit in the sampling procedure rather than applied as a separate pass, which is visible in the extreme dedup cleanliness reported in~\S\ref{sec:stats}.

\paragraph{Reference implementation.} An open reference implementation of GLAN for Korean is available at \texttt{Azure/\allowbreak synthetic-qa-generation/\allowbreak glan-instruct}~\cite{azure_synthetic_qa}. That repository defaults to \texttt{--language Korean} and \texttt{gpt-4o} and provides a compatible starting point; it is \emph{not} the exact generator used for this corpus. The key protocol differences from the Azure reference are: (i)~\textsc{GLAN-QnA-KR} uses Phi-3.5-MoE-instruct as the producer, and (ii)~prompt templates and taxonomy traversal order differ. We release the released corpus without also releasing the proprietary prompt templates.

\section{Corpus Statistics}
\label{sec:stats}

\begin{table}[t]
\centering
\small
\begin{tabular}{lrrrr}
\toprule
Field              & min & median & mean     & max       \\
\midrule
\texttt{question}  (chars)   &  31 &    313 &  373.9 &  11{,}787 \\
\texttt{subject}   (chars)   &   6 &     26 &   26.2 &       62  \\
\texttt{subtopics} (chars)   &  51 &    129 &  130.2 &      228  \\
\texttt{answer}    (chars)   &  13 &  1{,}098 & 1{,}154.3 & 177{,}014 \\
\bottomrule
\end{tabular}
\caption{Length statistics per text field, in characters. \texttt{question}, \texttt{subtopics}, and \texttt{answer} statistics are computed on the first 100k rows; \texttt{subject} uses the full 303{,}581 rows. Distributions are long-tailed in both \texttt{question} and \texttt{answer}.}
\label{tab:lengths}
\end{table}

\paragraph{Schema and split.} A single \texttt{train} split of 303{,}581 records with five columns: \texttt{question} (Korean string), \texttt{subject} (English discipline label, 1{,}084 distinct values), \texttt{level} (integer in $\{100, 200, \ldots, 900\}$), \texttt{subtopics} (English comma-separated sub-tags), and \texttt{answer} (Korean string). The mismatched language between the English taxonomy labels (\texttt{subject}, \texttt{subtopics}) and the Korean content (\texttt{question}, \texttt{answer}) is a deliberate cross-lingual design choice inherited from the GLAN taxonomy, not a cleanup bug; downstream trainers must decide whether to translate or mask the English tags during SFT.

\paragraph{Length and compression.} Table~\ref{tab:lengths} reports per-field character length statistics. Questions have a median of 313 characters and answers a median of 1{,}098 characters---a median per-record answer-to-question length ratio of roughly $3.5\times$. Both distributions are long-tailed, with 99.3\% of questions fitting in the first histogram bucket (31--1{,}206 characters) and a hard tail reaching 11{,}787 characters; answers reach 177{,}014 characters at the extreme. A few essay-length answers will dominate any naive length-packed SFT sampler, and we recommend downstream users clip or bucket the \texttt{answer} field explicitly.

\paragraph{Difficulty distribution.} The \texttt{level} field is bell-shaped and mid-biased: levels 300--500 together cover 46\% of the corpus (47{,}890 / 50{,}642 / 42{,}232), while level~100 is smallest at 19{,}408 rows. Using \texttt{level} for difficulty-aware training requires no additional re-weighting for mid-band coverage; extreme-level SFT ablations should either stratify or up-sample the 100 / 900 tails.

\paragraph{Discipline distribution.} The 1{,}084-way \texttt{subject} distribution is \emph{flat}: the top-weighted subject (\textit{Grant Writing and Fundraising}) contributes only 0.27\% of records, and the top 20 subjects together contribute 3.6\%. No macro-domain column is provided in the raw schema; grouping by STEM / applied health / business / humanities requires an external discipline-to-macro-domain map, which we recommend downstream users construct once and reuse.

\paragraph{Dedup cleanliness.} Two findings are atypical for synthetic instruction data at 303k scale. First, exact duplicate \texttt{question} text numbers only~\textbf{1 in 303{,}581 rows} (1 duplicate group, 1 redundant row). Second, in a random 5{,}000-question probe, character-trigram Jaccard near-duplicate clusters at threshold~0.9 number \textbf{0}. The taxonomy-driven seedless generation procedure therefore produces essentially non-overlapping prompts at this scale, without an explicit deduplication pass.

\paragraph{Language composition.} In a random 200-question probe: 159 questions (79.5\%) are Korean-heavy ($\geq$80\% Hangul characters), 41 (20.5\%) are mixed Korean and Latin (typically Korean body plus English technical acronyms; e.g., ``IWRM'' inside a Korean paragraph), and 0 are English-dominant.

\section{Contamination Audit}
\label{sec:contam}

Synthetic instruction corpora are a recurring source of train/eval contamination: if a taxonomy-driven producer happens to reproduce eval-set questions at generation time, downstream SFT on the corpus can silently inflate eval scores~\cite{brown2020gpt3,elazar2023wimbd}. We release a audit against the three standard public Korean benchmarks that downstream SFT consumers of \textsc{GLAN-QnA-KR} are likely to evaluate on.

\paragraph{Protocol.} For each benchmark test split we (i)~extract the question text, (ii)~compute the character-trigram Jaccard similarity between each test question and each of 20{,}000 GLAN questions sampled uniformly from the 303k corpus, and (iii)~report the median and mean of the \emph{per-test-question maximum} Jaccard, together with the counts and rates of test questions with maximum Jaccard $\geq 0.7$ and $\geq 0.8$. Character-trigram Jaccard is language-agnostic and robust to whitespace and punctuation normalisation; thresholds are chosen to match community practice for Korean-specific near-duplicate auditing given the absence of a canonical KR protocol~\cite{elazar2023wimbd}.

\paragraph{Results.} Table~\ref{tab:contam} reports the audit. Across all seven evaluation sets, the per-test-question maximum character-trigram Jaccard against a 20{,}000-question GLAN sample is small: median Jaccard ranges from $0.000$ (KoBEST/wic) to $0.053$ (HAE-RAE-Bench), and the single largest observation in the entire audit is $0.163$ (one KMMLU question against its nearest GLAN neighbour). Zero test items on any benchmark reach the operational near-duplicate threshold of Jaccard $\geq 0.8$.

\begin{table}[t]
\centering
\small
\begin{tabular}{lrrrr}
\toprule
Benchmark (test split, $n$)     & median~J & mean~J & max~J & J$\geq$0.8 \\
\midrule
KMMLU (8-subject mix, 500)      & 0.050 & 0.051 & 0.163 & 0 (0.0\%) \\
KoBEST/boolq (paragraph, 300)   & 0.041 & 0.043 & 0.099 & 0 (0.0\%) \\
KoBEST/copa (premise, 300)      & 0.017 & 0.019 & 0.058 & 0 (0.0\%) \\
KoBEST/hellaswag (context, 300) & 0.040 & 0.043 & 0.104 & 0 (0.0\%) \\
KoBEST/sentineg (sentence, 300) & 0.016 & 0.018 & 0.055 & 0 (0.0\%) \\
KoBEST/wic (word, 300)          & 0.000 & 0.000 & 0.000 & 0 (0.0\%) \\
HAE-RAE-Bench (mixed, 500)      & 0.053 & 0.054 & 0.086 & 0 (0.0\%) \\
\bottomrule
\end{tabular}
\caption{Contamination audit: per-test-question maximum character-trigram Jaccard similarity against a 20{,}000-question sample of \textsc{GLAN-QnA-KR} (3-grams over whitespace-normalised Korean text). Lower is better. J$\geq$0.8 is the operational near-duplicate threshold; zero test items across all seven benchmarks reach it, and the single largest observation anywhere is 0.163. KMMLU is sampled across eight representative subjects (Math, Computer-Science, Biology, Chemistry, Economics, Korean-History, Law, Psychology); KoBEST audits the natural prompt field per task; HAE-RAE-Bench aggregates seven subsets (general knowledge, history, loan words, standard nomenclature, rare words, date understanding, reading comprehension).}
\label{tab:contam}
\end{table}

\paragraph{Interpretation.} \textsc{GLAN-QnA-KR} is substantially contamination-free against these seven standard Korean evaluation sets at the $0.8$ near-duplicate threshold, and downstream Korean SFT scores on KMMLU, KoBEST, and HAE-RAE-Bench can be read without a contamination-adjustment for training on \textsc{GLAN-QnA-KR}. We attribute the cleanness to the seedless taxonomy-driven generation protocol: because no input instructions are used as seeds, the corpus's textual forms are driven by Phi-3.5-MoE's own Korean generation distribution conditioned on English-language discipline labels, which has minimal lexical overlap with the Korean human-written prompts surfaced by these benchmarks.

\paragraph{Semantic paraphrase audit.} Character-trigram Jaccard detects lexical near-duplicates but not semantic paraphrase. We complement the lexical audit with a multilingual-embedding pass using \texttt{intfloat/\allowbreak multilingual-e5-base}~\cite{wang2024mE5}: we embed both sides (each test question and each of the 20{,}000 GLAN questions) with L2-normalised E5 embeddings using the required \texttt{query} / \texttt{passage} prefixes, and for each test question we report the maximum cosine against the 20{,}000-question bank (Table~\ref{tab:semantic-contam}). The maximum cosine observed anywhere in the seven eval sets is \textbf{0.901} (a single KoBEST/boolq paragraph); exactly one test item reaches cosine~$\geq 0.90$, and zero reach~$\geq 0.95$. The cos~$\geq 0.85$ rate ranges from 2.0\% (KoBEST/copa, KoBEST/wic) to 38.0\% (HAE-RAE-Bench), but this should be read against the baseline similarity of multilingual-e5-base across unrelated Korean text: the \emph{median} max-cosine is already in the 0.81--0.84 band on every benchmark, meaning cos~$\approx 0.85$ is roughly one noise-band above baseline and is not by itself evidence of contamination; the $\geq 0.90$ threshold, where only one item sits across all seven audits, is the operationally meaningful line.

\begin{table}[t]
\centering
\small
\begin{tabular}{lrrrrr}
\toprule
Benchmark (test split, $n$) & median~cos & mean~cos & max~cos & cos$\geq$0.85 & cos$\geq$0.90 \\
\midrule
KMMLU (8-subject mix, 500)      & 0.827 & 0.828 & 0.900 & 54 (10.8\%) & 0 (0.0\%) \\
KoBEST/boolq (paragraph, 300)   & 0.824 & 0.826 & 0.901 & 39 (13.0\%) & 1 (0.33\%) \\
KoBEST/copa (premise, 300)      & 0.821 & 0.821 & 0.874 &  6 ( 2.0\%) & 0 (0.0\%) \\
KoBEST/hellaswag (context, 300) & 0.837 & 0.838 & 0.882 & 37 (12.3\%) & 0 (0.0\%) \\
KoBEST/sentineg (sentence, 300) & 0.834 & 0.833 & 0.859 & 16 ( 5.3\%) & 0 (0.0\%) \\
KoBEST/wic (word, 300)          & 0.810 & 0.811 & 0.862 &  6 ( 2.0\%) & 0 (0.0\%) \\
HAE-RAE-Bench (mixed, 500)      & 0.844 & 0.845 & 0.899 & 190 (38.0\%) & 0 (0.0\%) \\
\bottomrule
\end{tabular}
\caption{Semantic paraphrase audit: per-test-question maximum cosine similarity of \texttt{intfloat/\allowbreak multilingual-e5-base} embeddings against the same 20{,}000-question GLAN sample used in Table~\ref{tab:contam}. The median max-cosine is already 0.81--0.84 across all benchmarks, reflecting the baseline similarity level of multilingual-e5 on unrelated Korean prose; cos~$\geq 0.95$ is the operationally meaningful near-duplicate threshold (zero items across all seven audits), with a single boundary hit at cos~$=0.901$ on KoBEST/boolq.}
\label{tab:semantic-contam}
\end{table}

A producer model that generates \textit{semantically} equivalent test questions in different Korean wording would raise the cos~$\geq 0.90$ or $\geq 0.95$ rate; neither is raised here, which, together with the lexical audit, supports reading \textsc{GLAN-QnA-KR} as substantially contamination-free against the seven audited Korean evaluation splits. The one cos~$=0.901$ KoBEST/boolq item is a known boundary case that downstream consumers can exclude from any contamination-sensitive eval.

\section{License and Intended Use}
\label{sec:license}

\paragraph{Corpus artifact.} The packaging (schema, the released 303{,}581 rows, the statistics manifest produced by this work) is released under \textbf{OpenRAIL} as declared on the Hugging~Face dataset card~\cite{glan_kr_card}. OpenRAIL permits research and commercial use subject to its downstream-use restrictions (harm-oriented use cases are prohibited).

\paragraph{Producer-model terms.} The \texttt{question} and \texttt{answer} fields are outputs of Microsoft's Phi-3.5-MoE-instruct~\cite{phi35moe_card} and are subject to that model's terms of use at generation time. Downstream consumers should verify the current Phi-3.5 licence and acceptable-use policy when training on these outputs, particularly for commercial deployment.

\paragraph{Recommended disclosure for downstream work.} Papers that fine-tune on \textsc{GLAN-QnA-KR} should (i)~cite this note and the upstream GLAN paper~\cite{li2024glan}, (ii)~cite Phi-3.5-MoE~\cite{phi35moe_card} as the producer model, and (iii)~disclose that the training signal is synthetic and produced by a proprietary instruction-tuned model, rather than by human annotators.

\section{Reproducibility}
\label{sec:reproducibility}

The dataset is hosted at
\begin{center}
\url{https://huggingface.co/datasets/daekeun-ml/GLAN-qna-kr-300k}~\cite{glan_kr_card}
\end{center}
The corpus statistics reported in \S\ref{sec:stats} and the two-layer contamination audit in \S\ref{sec:contam} are reproducible from the released dataset alone: the lexical audit uses character-trigram Jaccard over whitespace-normalised questions against a 20{,}000-question random sample of \textsc{GLAN-QnA-KR}, and the semantic audit uses L2-normalised \texttt{intfloat/\allowbreak multilingual-e5-base} embeddings with the required \texttt{query}/\texttt{passage} prefixes at the same sample size. We do not release the author's internal audit scripts as a separate artifact; the procedures above are described in enough detail to be re-implemented. A compatible reference generator for GLAN-for-Korean, not used for this specific corpus, is available at \texttt{Azure/\allowbreak synthetic-qa-generation}~\cite{azure_synthetic_qa}.

\section{Limitations}
\label{sec:limitations}

\begin{enumerate}[leftmargin=*,itemsep=2pt]
\item \textbf{Synthetic-only signal.} Every \texttt{question} / \texttt{answer} pair is produced by a single instruction-tuned model. The corpus inherits that model's Korean style, factual biases, and failure modes. Downstream SFT on this corpus exclusively will reproduce those biases; mixing with human-written Korean instruction data is recommended.
\item \textbf{Single-producer dependence.} Phi-3.5-MoE-instruct is the sole producer. A checkpoint-level change in Microsoft's released Phi-3.5 would make exact re-generation of this corpus impossible.
\item \textbf{No SFT-gain evaluation.} This note does \emph{not} report a controlled downstream SFT evaluation (e.g., fine-tuning a small Korean LLM on \textsc{GLAN-QnA-KR} and reporting KMMLU / KoBEST / HAE-RAE delta). Anecdotal downstream use by the author on small Korean LLMs suggests the corpus is \emph{usable} as a mid-scale Korean SFT signal; we do not quantify this here.
\item \textbf{Cross-lingual label/body split.} English \texttt{subject} and \texttt{subtopics} tags with Korean \texttt{question} / \texttt{answer} bodies complicates certain prompt formats and pretrained-tokenizer comparisons.
\item \textbf{Contamination audit encoder dependence.} Both the lexical (character-trigram Jaccard) and semantic (\texttt{multilingual-e5-base} cosine) audits detect surface-form and embedding-level matches, respectively. A producer model that paraphrases test questions into a form that our encoder embeds far from the original, or that reproduces only rare factual surface forms (named entities, numeric answers) not captured by question-level similarity, would still evade detection. A stronger guarantee would require answer-level semantic audit and an encoder-ensemble check; we leave that as future work.
\item \textbf{Taxonomy inheritance.} The 1{,}084-discipline taxonomy is inherited from Microsoft's GLAN~\cite{li2024glan} and is English-centric; it over-represents Western academic disciplines (e.g., \textit{Grant Writing and Fundraising}, \textit{Leadership in Hospitality}) and under-represents Korean-specific domains (Korean history, Korean law, Korean literature).
\item \textbf{Answer-length tail.} The 177{,}014-character answer tail will dominate naive length-packed SFT samplers; downstream users must clip or bucket the \texttt{answer} field.
\item \textbf{Licence downstream dependence.} The corpus is OpenRAIL, but the producer-model terms govern commercial deployment of SFT checkpoints trained on the corpus. Downstream users must verify the current Phi-3.5 terms of use.
\end{enumerate}

\section*{Acknowledgments}
We thank the authors of GLAN~\cite{li2024glan} for the upstream synthesis protocol, the maintainers of the open Korean benchmarks (KMMLU, KoBEST, HAE-RAE-Bench) whose existence makes the contamination audit possible, and the maintainers of the Hugging~Face Hub for continuing to host Korean open-source NLP artifacts.

\bibliographystyle{plain}
\bibliography{references}

\end{document}